\documentclass[runningheads]{llncs}
\usepackage[T1]{fontenc}
\usepackage{graphicx}
\usepackage{amsmath,amsfonts}
\usepackage{algorithmic}
\usepackage{array}
\usepackage[caption=false,font=normalsize,labelfont=sf,textfont=sf]{subfig}
\usepackage{textcomp}
\usepackage{stfloats}
\usepackage{url}
\usepackage{verbatim}

\usepackage{listings}
\usepackage[table,xcdraw]{xcolor}
\usepackage{capt-of}
\usepackage{multirow}
\usepackage{pifont}
\usepackage{booktabs}
\usepackage{etoolbox}
\usepackage{xspace}
\usepackage{orcidlink}
\hypersetup{hidelinks} 

\makeatletter
\patchcmd{\@makecaption}
  {\scshape}
  {}
  {}
  {}
\makeatother

\newcommand{\rb}[1]{\textcolor{black}{#1}}

\newcommand{\model}{\texttt{CT}\xspace}

\usepackage{balance}

\def\ie{\emph{i.e. }} 
\def\wrt{w.r.t. }

\usepackage[capitalize]{cleveref}
\crefname{section}{Sec.}{Secs.}
\Crefname{section}{Section}{Sections}
\Crefname{table}{Table}{Tables}
\crefname{table}{Tab.}{Tabs.}

\begin{document}
\title{Co-Teaching for Unsupervised Domain Expansion}

\author{Hailan Lin\orcidlink{0009-0004-0529-3196} \and 
Qijie Wei\orcidlink{0009-0008-6895-5870} \and 
Kaibin Tian\orcidlink{0000-0002-3291-1146} \and 
Ruixiang Zhao\orcidlink{0009-0008-9984-1841} \and 
Xirong Li\thanks{Corresponding author: Xirong Li (xirong@ruc.edu.cn)}
\orcidlink{0000-0002-0220-8310}
}
\authorrunning{H. Lin et al.}
%
\institute{Renmin University of China\\
\url{https://github.com/ruc-aimc-lab/co-teaching}
}

\maketitle

\begin{abstract}
Unsupervised Domain Adaptation (UDA) essentially trades a model's performance on a source domain for improving its performance on a target domain. To overcome this, Unsupervised Domain Expansion (UDE) has been introduced, which adapts the model to the target domain while preserving its performance in the source domain. In both UDA and UDE, a model tailored to a given domain is assumed to well handle samples from the given domain. 
We question the assumption by reporting the existence of \emph{cross-domain visual ambiguity}: Due to the unclear boundary between the two domains, samples from one domain can be visually close to the other domain. Such sorts of samples are typically in the minority in their host domain, so they tend to be overlooked by the domain-specific model, but can be better handled by a model from the other domain. 
We exploit this finding by proposing \emph{Co-Teaching} (CT), which is instantiated with knowledge distillation based CT (kdCT) plus mixup based CT (miCT). Specifically, kdCT leverages a dual-teacher architecture to enhance the student network's ability to handle cross-domain ambiguity. Meanwhile, miCT further enhances the generalization ability of the student. Extensive experiments on image classification and driving-scene segmentation show the viability of CT for UDE.


\end{abstract}

\section{Introduction}
Unsupervised Domain Adaptation (UDA), aiming to adapt a model trained on a \emph{labeled} source domain for an \emph{unlabeled} target domain with no need of re-labeling any data, is crucial for real-world applications. 
With novel UDA methods continuously developed \cite{Li_2024_CVPR,Zhang_2025_CVPR}, we witness an ever-growing performance on the target domain, as manifested on public datasets such as Office-Home \cite{officehome} 
for image classification and ACDC \cite{acdc} for driving scene segmentation. However, it has been documented recently that UDA in fact trades the model's classification performance on the source domain for improving its performance on the target domain \cite{tomm21kdde,yang2021generalized}. Such a finding is disturbing, as it suggests that one may have to deploy simultaneously two models to support both domains. Even putting the doubled deployment cost aside, how to swiftly switch between the models is nontrivial, as from which domain a test sample comes is unknown. Source-domain performance degeneration puts the real-world use of UDA into question.

To remedy the issue, GSFDA \cite{yang2021generalized}  resorts to continual learning to preserve a model's ability on the source domain while being adapted to the target domain. Since the method assumes zero availability of the source-domain samples, it has little chance to recover once degeneration occurs. Indeed, our evaluation shows that GSFDA also suffers source-domain performance loss. 

To explicitly quantify the issue, a variant of UDA termed Unsupervised Domain Expansion (UDE) has been developed \cite{tomm21kdde}. UDE has the same starting point as UDA, \ie a set of \emph{labeled} training samples from the source domain and a set of \emph{unlabeled} training samples from the target domain. The key difference is that UDE explicitly reports the source-domain performance and consequently the performance on an expanded domain covering the source and target domains. The KDDE method for UDE assumes that a model tailored to a given domain can well handle samples from the given domain \cite{tomm21kdde}. Accordingly, KDDE runs in two steps, where two domain-specific models are first trained for the source and target domains, respectively.  It then performs knowledge distillation, where the dark knowledge of the source-specific (target-specific) teacher is transferred to a student model via source-domain (target-domain) samples exclusively. 

\begin{figure}[!htbp]
  \centering
  \includegraphics[width=\columnwidth]{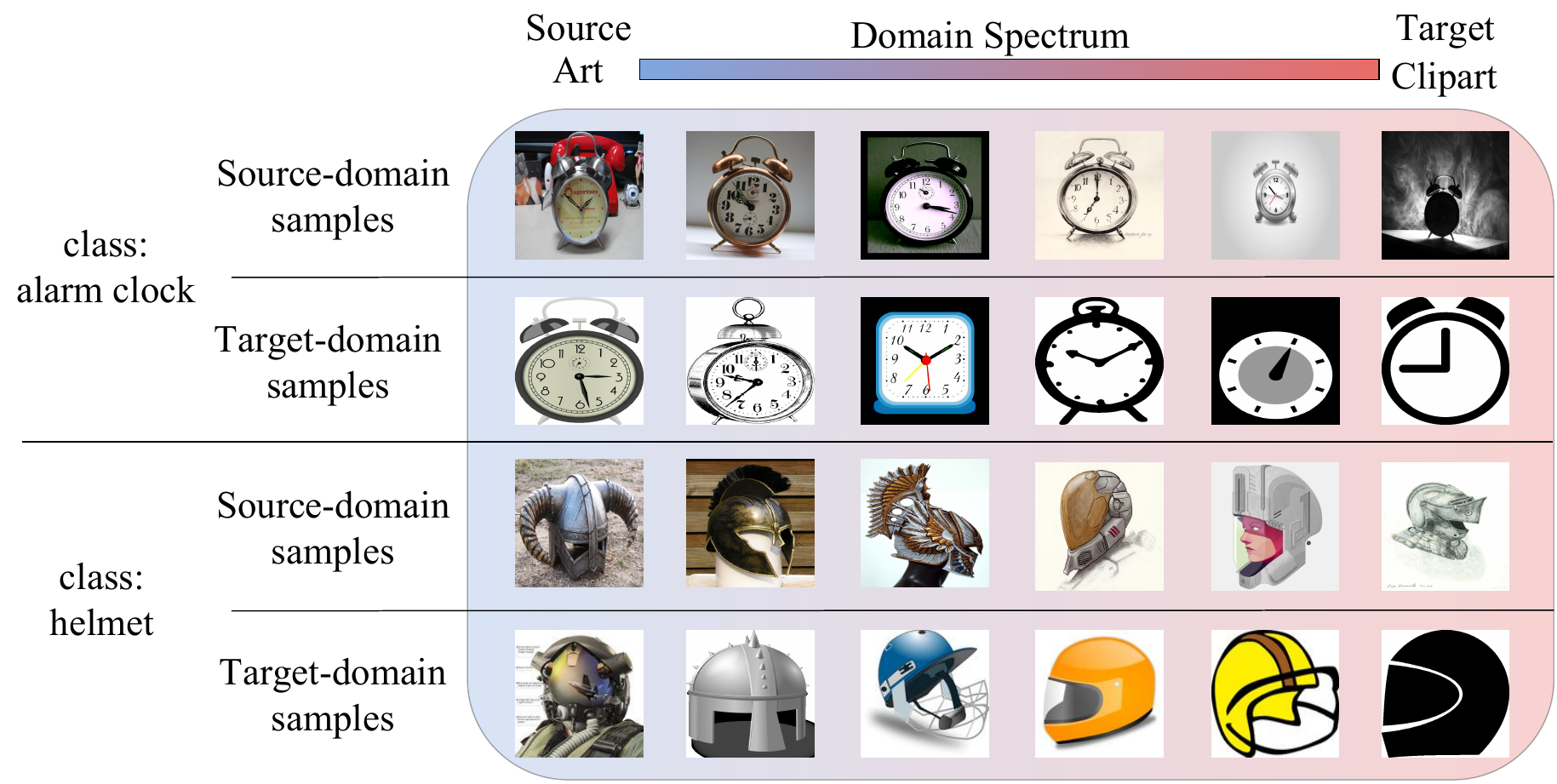}
  \caption{\textbf{Cross-domain visual ambiguity}. Samples from a target domain  (\emph{Clipart}) can be visually realistic as samples from a source domain (\emph{Art}), and vice versa. 
  }
  \label{fig:insight}
\end{figure}

We question the assumption of KDDE by reporting the existence of cross-domain visual ambiguity, see \cref{fig:insight}.
We consider a test image in domain $A$ \emph{cross-domain ambiguous} if the image is wrongly predicted by a model well-trained on domain $A$, yet correctly classified by a model targeted at a different domain $B$. \cref{tab:special} shows the percentage of such ambiguous images on Office-Home. These (minority) samples tend to be overlooked by domain-specific models.

\begin{table}[!htbp]
\caption{\textbf{Percentage of ambiguous test images on domain $A$}.}  
\centering
\scalebox{0.9}
{
\begin{tabular}{cl|cccc}

\hline
\multicolumn{2}{l|}{\multirow{2}{*}{\textbf{Dataset: Office-Home}}} & \multicolumn{4}{c}{\emph{Domain} $A$} \\ \cline{3-6} 
\multicolumn{2}{l|}{} & Art & Clipart & Product & Real \\ \hline
\multicolumn{1}{c|}{\multirow{4}{*}{\emph{Domain} $B$}} & Art & -- & 3.9 & 2.4 & 3.9 \\
\multicolumn{1}{c|}{} & Clipart & 5.0 & -- & 1.9 & 4.1 \\
\multicolumn{1}{c|}{} & Product & 6.0 & 4.0 & -- & 3.9 \\
\multicolumn{1}{c|}{} & Real & 7.5 & 4.6 & 2.6 & -- \\ 
\hline
\end{tabular}
}

\label{tab:special}
\end{table}

\begin{figure}[!htbp]
  \centering
  \includegraphics[width=1\columnwidth]{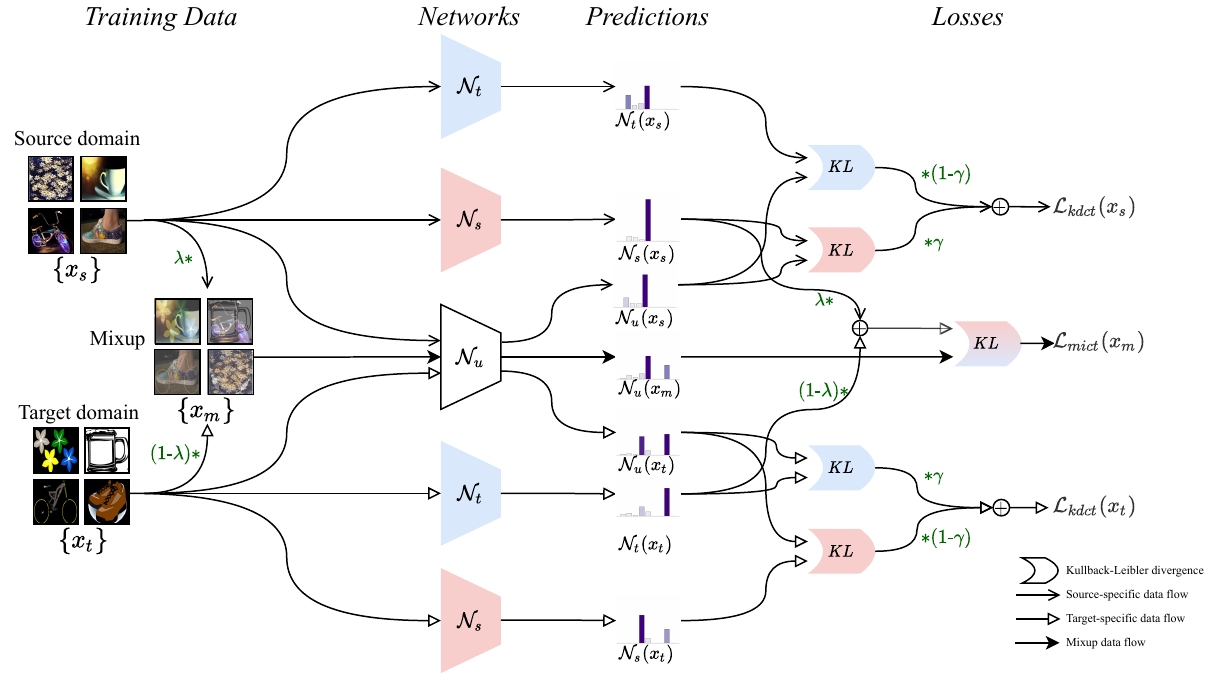}
  \caption{\textbf{Proposed \model method}. Given labeled data $\{(x_s, y_s)\}$ from a source domain and unlabeled data $\{x_t\}$ from a target domain, CT obtains a 
  \emph{domain-expanded} network $\mathcal{N}_u$ by two-stage training. In the first stage, two \emph{domain-specific} teacher networks $\mathcal{N}_s$ and $\mathcal{N}_t$ are obtained, where $\mathcal{N}_s$ for the source domain is trained on $\{(x_s,y_s)\}$ by standard supervised learning, whilst $\mathcal{N}_t$ for the target domain is trained on $\{(x_s,y_s)\}$ and $\{x_t\}$ by an existing UDA method. 
  {In the second stage, $\mathcal{N}_s$ and $\mathcal{N}_t$ co-teach $\mathcal{N}_u$ via}
  \textbf{knowledge distillation based CT} (kdCT) that minimizes $L_{kdct}(\{x_s\}) + L_{kdct}(\{x_t\})$, 
  and \textbf{mixup based CT} (miCT),  minimizing  $L_{mict}(\{x_m\})$ for mixup instances $\{x_m\}$.
  Once trained, only $\mathcal{N}_u$ is needed for inference. 
  }
  \label{fig:overview}
\end{figure}

In order to tackle such ambiguity, we propose Co-Teaching (\model) for UDE. 
The proposed method comprises knowledge distillation based CT (kdCT) and mixup based CT (miCT), see \cref{fig:overview}. Specifically, kdCT transfers knowledge from a leader-teacher network and an assistant-teacher network to a student network for cross-domain ambiguity, while miCT further enhances the generalization ability of the student. Extensive experiments on multi-class image classification and driving scene segmentation verify the effectiveness of \model.

\section{Related Work}

\textbf{Progress on UDA}.
The major line of research on UDA is to learn domain-invariant feature representations, 
either by domain discrepancy reduction  \cite{ddc,CHEN2018270,kang2019contrastive} or by adversarial training \cite{jmlr16-dann,nips18-cdan,CHEN2020107440}. 
\rb{More recently, the mixup technique, originally developed for supervised deep learning \cite{zhang2018mixup}, has been actively leveraged for learning domain-invariant features \cite{dmada,na2021fixbi}.}
FixBi, for instance, uses fixed ratio-based mixup to train a source-biased classifier and a target-biased classifier \cite{na2021fixbi}, enforcing consistency on domain-mixed samples. 
More recent studies exploit pretrained vision-language 
models  \cite{du2024domain,Li_2024_CVPR}.

There is also an increasing interest in extending image-level UDA to the pixel level for cross-domain semantic image segmentation. AdaSegNet \cite{tsai2018learning} and Advent \cite{Vu_2019_CVPR} use adversarial training at the output space, while pixel-level adaptation is performed in DCAN \cite{wu2018dcan} and Cycada \cite{hoffman2018cycada}.
FDA \cite{yang2020fda} uses self-predicted labels for self-supervised training.

Our proposed \model conceptually differs from the above works as it essentially performs meta learning on top of a specific UDA method. Moreover, in contrast to the prior art which cares only the target-domain performance, CT aims for a broader scope covering both the source and target domains. 
{While there are efforts on preventing the source-domain performance deterioration \cite{yang2021generalized,rostami2021lifelong,lv2021pareto}, our experiment indicates that the deterioration remains.}

\textbf{Progress on UDE}.
\model is inline with KDDE \cite{tomm21kdde}, as both aim for training a model that suits the expanded domain by knowledge distillation (KD). 
{In KDDE, however, knowledge from the source-domain teacher is only transferred via source samples, and similarly for the target domain. Consequently, KDDE lacks the ability to leverage the teacher network derived from one domain in handling cross-domain ambiguity for the other domain. While multi-teacher KD has been used in multi-source UDA by training domain-adapted teachers through target and multiple source domain pairings \cite{LIU2022108430,LUO2022108955}, this tactic is inapplicable to the single-source scenario as considered in this work. 
}

\textbf{Co-Teaching}. The term co-teaching has been used in other contexts which conceptually and technically differ from ours. For supervised learning, Decoupling \cite{malach2017decoupling} and Co-Teach \cite{han2018co} aim for label denoising within a single domain. Decoupling simultaneously trains two models $h_1$ and $h_2$, updating them with samples having $h_1(x) \neq h_2(x)$ in a given mini-batch. Co-Teach alternately uses samples correctly classified by one model to train the other model. Since both methods are fully supervised, they are inapplicable for UDA / UDE.
CGCT \cite{roy2021curriculum} proposes a co-teaching strategy using a dual-head classifier to provide pseudo labels for unlabeled target-domain samples. CT is a \emph{model-agnostic} meta learner, so any UDA method including CGCT can in principle be used as its UDA module.

\section{Method}
\subsection{Problem Formalization}

We use $x$ to indicate a specific sample. For a manually labeled sample, we use $y$ to indicate its label, which can be a one-hot vector or a multi-dimensional binary mask. Let $\mathcal{N}$ be a deep neural network, which outputs $\mathcal{N}(x)$ that well matches the (unknown) label of a novel sample. Following \cite{tomm21kdde}, we formalize the UDE task as follows. Given a set of $n_s$ \emph{labeled} samples $\{(x_s,y_s)\}$ randomly sampled from a source domain $D_s$ and a set of $n_t$ \emph{unlabeled} samples $\{x_t\}$ randomly sampled from a target domain $D_t$, the goal is to train $\mathcal{N}$ that works for an expanded domain covering both $D_s$ and $D_t$ which named as $D_{s+t}$. 

The previous approach to UDE is KDDE \cite{tomm21kdde}. At a high level, KDDE works in two stages. In the first stage, two domain-specific teacher networks $\mathcal{N}_s$ and $\mathcal{N}_t$ are trained, where $\mathcal{N}_s$ for $D_s$ is learned from the labeled set $\{(x_s, y_s)\}$ by standard supervised learning, while $\mathcal{N}_t$ for $D_t$ is trained on $\{(x_s, y_s)\}$ and $\{x_t\}$ by an off-the-shelf UDA method. In the second stage, knowledge distillation (KD) is performed to inject the dark knowledge of the teacher networks into a student network $\mathcal{N}_u$, which will be eventually used for inference. Depending on the domain identity of a training sample, KDDE uses the two teachers, \ie $\mathcal{N}_s$ to deal with samples from $D_s$ and $\mathcal{N}_t$ for samples from $D_t$, see \cref{eq:kdde}. 
\begin{small}
\begin{equation} \label{eq:kdde}
\left\{
\begin{array}{ll}
\mathcal{N}_s  & \leftarrow \mbox{supervised-learning}(\{(x_s,y_s)\}) \\
\mathcal{N}_t  & \leftarrow \mbox{UDA}(\{(x_s,y_s)\}, \{x_t\}) \\
\mathcal{N}_u &  \leftarrow
\left\{
\begin{array}{l}
\mbox{KD}(\mathcal{N}_s,\{x\}), \quad x\in D_s  \\
\mbox{KD}(\mathcal{N}_t,\{x\}), \quad x\in D_t
\end{array} 
\right.
\end{array} 
\right.
\end{equation}
\end{small}

The two-stage property of KDDE ensures flexibility in choosing UDA methods for implementing $\mathcal{N}_t$. We inherit this property and introduce a novel Co-Teaching (\model) method into the second stage. 

\subsection{Framework}
\label{subsec:method_stage2}

\model consists of knowledge distillation based CT (kdCT) and mixup based CT (miCT).
As \cref{fig:overview} shows, kdCT allows the student network to simultaneously learn the two teacher networks' dark knowledge about every training sample. 
Meanwhile, miCT improves the generalization ability of the student by using the mixup technique in a cross-domain manner. As the two implementations of \model are orthogonal to each other, they can be used either alone or jointly. 

\subsubsection{Knowledge Distillation based Co-Teaching}

We depart from a standard KD process with one student network $\mathcal{N}_u$ and one teacher network, either $\mathcal{N}_s$ or $\mathcal{N}_t$. Let us consider $\mathcal{N}_s$ for instance. 
Given a set of samples $\{x\}$, KD from $\mathcal{N}_s$ to $\mathcal{N}_u$ is achieved by minimizing the Kullback-Leibler (KL) divergence between $\mathcal{N}_s(\{x\})$ and $\mathcal{N}_u(\{x\})$, \rb{denoted} as $KL(\mathcal{N}_s(\{x\}),\mathcal{N}_u(\{x\}))$. \rb{Similarly}, we have the loss of KD from $\mathcal{N}_t$ to $\mathcal{N}_u$ as $KL(\mathcal{N}_t(\{x\}),\mathcal{N}_u(\{x\}))$. 
{For multi-teacher KD, simply averaging losses is suboptimal since teachers specialize in distinct domains. Therefore, we shall not treat them equally. For training samples from $D_s$, we expect that $\mathcal{N}_s$ leads the teaching process, while the $\mathcal{N}_t$ acts as an assistant, and vice versa. This is implemented via a parameter $\gamma$ that weighs the importance of each teacher
in the kdCT process:}

\begin{equation} \label{eq:kdct}
\resizebox{0.7\hsize}{!}{
    $ 
    L_{kdct}(\{x\})= 
    \left\{
    \begin{array}{l}
        \gamma \cdot KL(\mathcal{N}_{s},\mathcal{N}_u)+ 
        (1-\gamma) \cdot  KL(\mathcal{N}_{t},\mathcal{N}_u) \quad x \in D_s
        \\
        \\
        \gamma \cdot KL(\mathcal{N}_{t},\mathcal{N}_u)+ 
        (1-\gamma) \cdot  KL(\mathcal{N}_{s},\mathcal{N}_u) \quad x \in D_t
    \end{array} \right.
    $ 
}
\end{equation}

{Given \cref{eq:kdct}, KDDE can now be viewed as a special case of kdCT with $\gamma=1$. kdCT extends the loss of KDDE
to exploit multiple teachers in a biased manner, which improves kdCT in making correct decisions on samples of domain ambiguity. We define the overall loss $L_{kdct}$ as the sum of two domain-specific losses $L_{kdct}(\{x_s\})$ and $L_{kdct}(\{x_t\})$ by \cref{eq:kdct}.}

{We assign $\gamma$ larger than $0.5$ to emphasize the leading-teacher network, \ie $\mathcal{N}_s$ for samples from $D_s$ and $\mathcal{N}_t$ for samples from $D_t$. To enhance robustness against noise, we introduce randomness by sampling $\gamma$ per mini-batch from $Beta(\alpha,\beta)$, whose shape parameters enable diversified probability distributions\cite{fawzi2016robustness,he2019parametric}.}

\subsubsection{Mixup based Co-Teaching}

The mixup technique \cite{zhang2018mixup}, synthesizing a new training sample by a convex combination of two real samples, is shown to be effective for improving image classification networks. We thus re-purpose this technique to generate new domain-expanded samples denoted by $\{x_m\}$. In particular, $x_m$ is obtained by blending $x_s$ randomly chosen from $D_s$ with $x_t$ randomly chosen from $D_t$. 

Our mixup based \model (miCT) is implemented by transferring the two-teacher knowledge via mixed samples to the student network. The teachers' joint knowledge \wrt $x_m$ is reflected by combined prediction denoted as $\hat{y}_m$. Given $\lambda  \sim Beta(1,1)$ as a mixup rate, the loss of miCT $L_{mict}$ is computed as
\begin{equation} \label{eq:mict}
\left\{ \begin{array}{ll}
x_m & = \lambda \cdot x_s + (1-\lambda) \cdot x_t \\
\hat{y}_m &= \lambda  \cdot \mathcal{N}_s(\{x_s\})+(1-\lambda) \cdot \mathcal{N}_t(\{x_t\}) \\
L_{mict} & =KL(\hat{y}_m , \mathcal{N}_u(\{x_m\}))
\end{array}\right.
\end{equation}
Both $L_{kdct}$ and $L_{mict}$ are KL-divergence based losses for knowledge distillation. So they can be directly summed up and minimized together for the joint use of kdCT and miCT.

\section{Experiments}

We evaluate \model for multi-class image classification and driving scene segmentation.
We use ResNet-50 \cite{he2016deep} as the student network for image classification, and DeepLabv2 \cite{chen2017deeplab} for semantic segmentation, unless otherwise stated.
It is worth pointing out that UDE as an emerging topic is less studied. 
So for a fair and comprehensive evaluation, we organize the baselines into the following three groups: methods targeted at UDE, methods targeted at UDA,  and methods technically related. All experiments are run with PyTorch on two NVIDIA Tesla P40 cards. 

\subsection{Task 1. Multi-Class Image Classification} \label{ssec:exp-img-clf}

\subsubsection{Experimental Setup} \label{sssec:img-clf-setup} 
We adopt the popular Office-Home dataset \cite{officehome},  which contains 15,588 images of 65 object classes common in office and home scenes with four different domains, \ie artistic images (A), clip art (C), product images (P), and real-world images (R). 
We adopt the data split of \cite{tomm21kdde}: images per domain have been divided randomly into training and testing subsets\footnote{\url{https://github.com/li-xirong/ude}}. 
Pairing the individual domains leads to 12 UDE tasks in total.

\textbf{Baseline methods}. 
We include as a baseline ResNet-50 trained by standard supervised learning on $D_s$. 
As mentioned above, we compare with existing methods from the following three groups: \\
$\bullet$ Method for UDE: 
KDDE \cite{tomm21kdde}. \\
$\bullet$ Methods for  UDA: 
DDC \cite{ddc}, 
DANN \cite{jmlr16-dann}, 
DAAN \cite{icdm19-daan}, 
CDAN \cite{nips18-cdan}, 
SRD \cite{tang2020unsupervised}, 
PDA \cite{lv2021pareto},
GSFDA \cite{yang2021generalized}, 
CGCT \cite{roy2021curriculum},
FixBi \cite{na2021fixbi}, SDAT \cite{rangwani2022closer} and ELS \cite{zhang2023free}.
\\
$\bullet$  Method technically related: MultiT \cite{you2017learning}. \\
Since each method is trained on the same data to yield a distinct ResNet-50 model for inference, our setup allows for a fair, head-to-head comparison.

In order to study whether CT also works with a transformer-based UDA method, we try CDTrans \cite{xu2022cdtrans}. Different from the baselines mentioned above, CDTrans uses DeiT-Base \cite{touvron2021training} as its backbone. We simply use the same training protocol (optimizer, initial learning rate, learning rate adjustment strategy, \emph{etc.}) as used for CNN.

\textbf{Performance metric}. We report accuracy (\%), \ie the percentage of test images correctly classified. 

\subsubsection{Results}

\begin{table}[!htbp]
\centering
 \renewcommand\arraystretch{1.1}
\caption{\textbf{Multi-class image classification}. Methods per group are sorted in terms of their expanded-domain performance. Top performers within each group are highlighted with bold font.}
\label{tab:total}
\scalebox{0.9}{
\begin{tabular}{@{}lrrr@{}}
\toprule
\textbf{Method} & \textbf{Source} $D_s$ & \textbf{Target} $D_t$ & \textbf{Expanded} $D_{s+t}$ \\
\midrule
ResNet-50 as $\mathcal{N}_s$  & 82.43 & 57.84 & 70.13 \\ 
\midrule
\textit{Choice of}~$\mathcal{N}_t$: &  &   & \\ 
CDAN \cite{nips18-cdan} & 80.36 & 61.57 & 70.96  \\
DANN \cite{jmlr16-dann} & 81.36 & 60.65 & 71.01 \\
CGCT \cite{roy2021curriculum} & 79.70 & 61.44 & 70.57  \\
FixBi \cite{na2021fixbi} & 77.10 & 64.31 & 70.71 \\
DDC \cite{ddc} & 82.35 & 60.51 & 71.43  \\
DAAN \cite{icdm19-daan} & \textbf{82.38} & 60.84 & 71.62 \\
SRDC \cite{tang2020unsupervised} & 78.68 & 65.30 & 71.99 \\ 
GSFDA \cite{yang2021generalized} & 79.90 & 66.53 & 73.22 \\
SDAT \cite{rangwani2022closer} &  81.48 & 68.42 & 74.95 \\
ELS \cite{zhang2023free} & 81.74 & \textbf{68.51} & \textbf{75.13} \\
\midrule
DDC as $\mathcal{N}_t$: &  &   &  \\
PDA \cite{lv2021pareto} & 76.90 & 54.01 & 65.46 \\
KDDE \cite{tomm21kdde} & 82.74 & 62.19 & 72.47  \\
\model & \textbf{82.92} & \textbf{63.06} & \textbf{72.99} \\ 
\midrule
CDAN as $\mathcal{N}_t$: &  &   &  \\
PDA & 78.44 & 57.65 & 68.04  \\
KDDE  & 81.03 & 62.96 & 72.00   \\
\model & \textbf{82.17} & \textbf{64.55} & \textbf{73.36} \\ 
\midrule
\textit{SRDC as} $\mathcal{N}_t$: &  &   &  \\
MultiT \cite{you2017learning} & 82.23 & 61.66 & 71.94 \\
KDDE  & 81.54 & 67.20 & 74.37  \\ 
\model  & \textbf{82.32} & \textbf{67.45} & \textbf{74.89} \\ 
\midrule
\model(FixBi as $\mathcal{N}_t$)  &  80.88 & 65.46 &  73.17 \\
\model(SDAT as $\mathcal{N}_t$)  &  81.87 & \textbf{68.86} & 75.37 \\
\model(ELS as $\mathcal{N}_t$) & \textbf{82.10} & 68.82 & \textbf{75.46} \\
\midrule
\midrule
DeiT-Base \cite{touvron2021training} as $\mathcal{N}_s$ & \textbf{88.31} & 72.38 & 80.35 \\
CDTrans \cite{xu2022cdtrans} as $\mathcal{N}_t$ & 85.37 & 78.78 & 82.07  \\
\model  & 88.04 & \textbf{79.19} & \textbf{83.62} \\
\bottomrule
\end{tabular}
}
\end{table}

\cref{tab:total} shows the performance of the varied methods on the source ($D_s$), target ($D_t$) and expanded ($D_{s+t}$) domains, respectively.
The UDA methods consistently show performance degeneration on $D_s$, including GSFDA (from 82.43 to 79.90) which aims for maintaining the source-domain performance.
For the UDA setting wherein only the target-domain performance matters, \model compares favorably against the best UDA baseline (ELS).
As for UDE, \model is again the best. The lower performance of MultiT than KDDE and \model confirms our hypothesis that the two teacher networks shall not be treated equally in the knowledge distillation process. 
As shown in the last three rows of \cref{tab:total}, 
\model also works with the Transformer-based UDA method (CDTrans).

\begin{figure}[!hthbp]
  \centering
  \includegraphics[width=0.9\columnwidth]{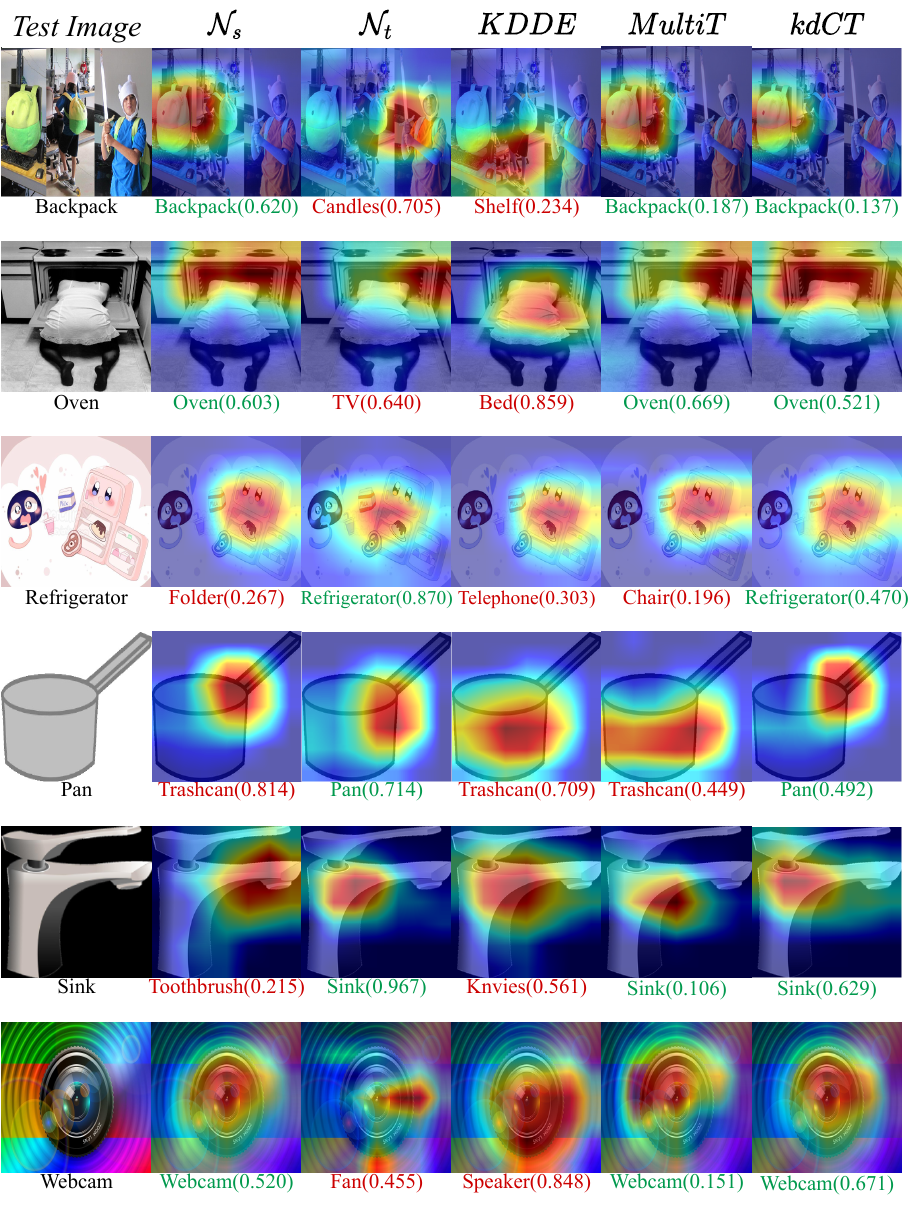}
  \caption{
  \textbf{Grad-CAM \cite{gradcam} visualization}.
  The top three rows are from a source domain (\emph{Art}), while the bottom three rows are from a target domain (\emph{Clipart}). Texts under heatmaps are predicted labels with scores.
  }
  \label{fig:cam}
\end{figure}

\textbf{On addressing the cross-domain ambiguity}. 
A fine-grained analysis is shown in \cref{tab:ambiguity}. 
We can mostly attribute the success of \model to its superior performance on the inconsistent group, which confirms the effectiveness of \model's compensation mechanism. 
As \cref{fig:cam} shows, the activated regions produced by the proposed method are more precise than the others. 
Both quantitative and qualitative results justify the efficacy of \model.
\begin{table}[!t]
\centering
\renewcommand\arraystretch{1.1}
\caption{\textbf{Fine-grained analysis}.  Each test set is divided into two disjoint subsets, \ie consistent and inconsistent, where each sample $x$ in the consistent set has $\mathcal{N}_s(x) == \mathcal{N}_t(x)$, while each sample in the inconsistent set has $\mathcal{N}_s(x) \neq \mathcal{N}_t(x)$. The classification accuracy score is calculated per subset. The gain of \model against KDDE and MultiT is  mostly attributed to the method's better performance on the inconsistent group.}
\scalebox{0.95}
{
\begin{tabular}{@{}lcc|cc|cc@{}}
\toprule
\multirow{2}{*}{\begin{tabular}[c]{@{}c@{}} \textbf{Task} \\ R$\rightarrow$P \end{tabular}} & \multicolumn{2}{c|}{\begin{tabular}[c]{@{}c@{}} \textit{$D_s$} \end{tabular}} & \multicolumn{2}{c|}{\begin{tabular}[c]{@{}c@{}} \textit{$D_t$} \end{tabular}} & \multicolumn{2}{c}{\begin{tabular}[c]{@{}c@{}} \textit{$D_{s+t}$} \end{tabular}} \\ \cline{2-7} 
 & \cellcolor[HTML]{DBF5DB}{\begin{tabular}[c]{@{}c@{}}$=$\end{tabular}} & \cellcolor[HTML]{F8CECC}{\begin{tabular}[c]{@{}c@{}}$\neq$
 \end{tabular}} & \cellcolor[HTML]{DBF5DB}{\begin{tabular}[c]{@{}c@{}}$=$\end{tabular}} & \cellcolor[HTML]{F8CECC}{\begin{tabular}[c]{@{}c@{}}$\neq$\end{tabular}} & \cellcolor[HTML]{DBF5DB}{\begin{tabular}[c]{@{}c@{}}$=$\end{tabular}} & \cellcolor[HTML]{F8CECC}{\begin{tabular}[c]{@{}c@{}}$\neq$\end{tabular}} \\ \midrule
$\mathcal{N}_{s}$ & \cellcolor[HTML]{DBF5DB}{90.00} & \cellcolor[HTML]{F8CECC}{42.75} & \cellcolor[HTML]{DBF5DB}{87.54} & \cellcolor[HTML]{F8CECC}{21.88} & \cellcolor[HTML]{DBF5DB}{88.80} & \cellcolor[HTML]{F8CECC}{30.20} \\
$\mathcal{N}_{t}$ & \cellcolor[HTML]{DBF5DB}{90.00} & \cellcolor[HTML]{F8CECC}{30.07} & \cellcolor[HTML]{DBF5DB}{87.54} & \cellcolor[HTML]{F8CECC}{48.31} & \cellcolor[HTML]{DBF5DB}{88.80} & \cellcolor[HTML]{F8CECC}{41.04} \\
KDDE & \cellcolor[HTML]{DBF5DB}{88.96} & \cellcolor[HTML]{F8CECC}{42.39} & \cellcolor[HTML]{DBF5DB}{87.32} & \cellcolor[HTML]{F8CECC}{51.20} & \cellcolor[HTML]{DBF5DB}{88.16} & \cellcolor[HTML]{F8CECC}{47.69} \\
MultiT & \cellcolor[HTML]{DBF5DB}{88.96} & \cellcolor[HTML]{F8CECC}{44.57} & \cellcolor[HTML]{DBF5DB}{86.61} & \cellcolor[HTML]{F8CECC}{41.59} & \cellcolor[HTML]{DBF5DB}{87.81} & \cellcolor[HTML]{F8CECC}{42.77} \\
\model & \cellcolor[HTML]{DBF5DB}{89.53} & \cellcolor[HTML]{F8CECC}{47.46} & \cellcolor[HTML]{DBF5DB}{87.54} & \cellcolor[HTML]{F8CECC}{52.88} & \cellcolor[HTML]{DBF5DB}{88.56} & \cellcolor[HTML]{F8CECC}{50.72} \\ 
\midrule
\multirow{2}{*}{\begin{tabular}[c]{@{}c@{}} \textbf{Task} \\ A$\rightarrow$C\end{tabular}} & \multicolumn{2}{c}{\begin{tabular}[c]{@{}c@{}}\textit{$D_s$}\end{tabular}} & \multicolumn{2}{c}{\begin{tabular}[c]{@{}c@{}}\textit{$D_t$}\end{tabular}} & \multicolumn{2}{c}{\begin{tabular}[c]{@{}c@{}}\textit{$D_{s+t}$}\end{tabular}} \\ \cline{2-7} 
 & \cellcolor[HTML]{DBF5DB}{\begin{tabular}[c]{@{}c@{}}$=$\end{tabular}} & \cellcolor[HTML]{F8CECC}{\begin{tabular}[c]{@{}c@{}}$\neq$\end{tabular}} & \cellcolor[HTML]{DBF5DB}{\begin{tabular}[c]{@{}c@{}}$=$\end{tabular}} & \cellcolor[HTML]{F8CECC}{\begin{tabular}[c]{@{}c@{}}$\neq$\end{tabular}} & \cellcolor[HTML]{DBF5DB}{\begin{tabular}[c]{@{}c@{}}$=$\end{tabular}} & \cellcolor[HTML]{F8CECC}{\begin{tabular}[c]{@{}c@{}}$\neq$\end{tabular}} \\ \midrule
$\mathcal{N}_{s}$ & \cellcolor[HTML]{DBF5DB}{85.19} & \cellcolor[HTML]{F8CECC}{43.30} & \cellcolor[HTML]{DBF5DB}{66.50} & \cellcolor[HTML]{F8CECC}{16.13} & \cellcolor[HTML]{DBF5DB}{74.41} & \cellcolor[HTML]{F8CECC}{22.90} \\
$\mathcal{N}_{t}$ & \cellcolor[HTML]{DBF5DB}{85.19} & \cellcolor[HTML]{F8CECC}{18.30} & \cellcolor[HTML]{DBF5DB}{66.50} & \cellcolor[HTML]{F8CECC}{27.61} & \cellcolor[HTML]{DBF5DB}{74.41} & \cellcolor[HTML]{F8CECC}{25.31} \\
KDDE & \cellcolor[HTML]{DBF5DB}{81.90} & \cellcolor[HTML]{F8CECC}{36.14} & \cellcolor[HTML]{DBF5DB}{65.13} & \cellcolor[HTML]{F8CECC}{29.99} & \cellcolor[HTML]{DBF5DB}{72.26} & \cellcolor[HTML]{F8CECC}{31.52} \\
MultiT & \cellcolor[HTML]{DBF5DB}{83.87} & \cellcolor[HTML]{F8CECC}{40.19} & \cellcolor[HTML]{DBF5DB}{65.45} & \cellcolor[HTML]{F8CECC}{27.51} & \cellcolor[HTML]{DBF5DB}{73.25} & \cellcolor[HTML]{F8CECC}{30.67} \\
\model & \cellcolor[HTML]{DBF5DB}{82.98} & \cellcolor[HTML]{F8CECC}{43.30} & \cellcolor[HTML]{DBF5DB}{66.75} & \cellcolor[HTML]{F8CECC}{32.78} & \cellcolor[HTML]{DBF5DB}{73.62} & \cellcolor[HTML]{F8CECC}{35.40} \\ 
\bottomrule
\end{tabular}
}

\label{tab:ambiguity}
\end{table}

\begin{table}[!htbp]
\centering
\renewcommand\arraystretch{1.1}
\caption{\textbf{Ablation Study}.}
\scalebox{0.95}
{
\begin{tabular}{@{}ccrrr@{}}
\toprule
\multicolumn{1}{c}{\textbf{kdCT}} & \textbf{miCT} & $D_s$ &  $D_t$ &  $D_{s+t}$ \\ 
\midrule
\multicolumn{2}{c}{DDC as $\mathcal{N}_t$} &   &   &   \\
 \checkmark &  & 82.85 & 62.42 & 72.63 \\
 &  \checkmark & 80.32 & 61.90 & 71.11 \\
 \checkmark &  \checkmark & \textbf{82.92} & \textbf{63.06} & \textbf{72.99} \\  
 \midrule

\multicolumn{2}{c}{SRDC as $\mathcal{N}_t$} &  &  &  \\ 
 \checkmark &  & \textbf{82.52} & 67.19 & 74.86 \\
 & \checkmark & 77.46 & 63.85 & 70.65 \\
 \checkmark &  \checkmark& 82.32 & \textbf{67.45} & \textbf{74.89} \\
\bottomrule
\end{tabular}
}

\label{tab:ablation}
\end{table}

\textbf{Influence of kdCT and miCT}. As   \cref{tab:ablation} shows, kdCT is better than miCT when used alone. The lower performance of miCT is because we train on mixup samples exclusively, without using original samples. 
The joint use of kdCT and miCT is recommended for image classification.

\begin{table}[!t]
\centering
\renewcommand\arraystretch{1.1}
\caption{\textbf{Effect of $\gamma$ on kdCT}.} 

\label{tab:choosebeta}
\scalebox{0.9}
{
\begin{tabular}{@{}lrrr@{}}
\hline
 & $D_s$ & $D_t$ & $D_{s+t}$ \\ \hline
\multicolumn{4}{l}{$(\alpha,\beta)$ of the beta distribution}  \\
10, 1 & 82.85 & {\color[HTML]{FE0000}\textbf{62.42}} & {\color[HTML]{FE0000}\textbf{72.63}} \\
5, 1 & 82.84 & 62.02 & 72.43 \\
1, 1 & {\color[HTML]{FE0000}\textbf{82.86}} & 61.54 & 72.20 \\
1, 5 & 82.78 & 60.61 & 71.70 \\
1, 10 & 82.65 & 60.61& 71.63 \\
\hline
\textit{Fixed} & & &  \\
0.5 & 82.75 &	61.41 & 72.08\\
0.909 & 82.81 & 61.67 & 72.24  \\
1 & 82.74 &	62.19 & 72.47 \\
\hline
\end{tabular}
}

\end{table}

\textbf{Effect of $\gamma$}.  \cref{tab:choosebeta} shows the performance of kdCT given  $\gamma$ specified in varied manners. 
We observe that using the fixed value $0.909$, which is the expectation value of $Beta(10,1)$, results in lower performance, justifying the benefit of using $\gamma$ in a stochastic manner.

\subsection{Task 2. Driving Scene Segmentation} 

\subsubsection{Experimental Setup}
We follow \cite{acdc}, using Cityscapes \cite{cityscapes} as $D_s$ and ACDC \cite{acdc} as $D_t$. Both datasets have pixel-level ground truth.
Different from Cityscapes consisting of normal lighttime driving scenes, ACDC has four adverse conditions, 
see Fig. \ref{fig:segmentation}. 
We adopt their official data splits, \ie 2,975 training and 500 test images in Cityscapes and 1,600 training and 406 test images in ACDC. Previous work on semantic image segmentation \cite{panfilov2019improving} reports that the mixup technique has an adverse effect, which is also observed in our preliminary experiment on driving scene segmentation. We therefore use kdCT for this task.

\textbf{Baselines}. We again compare with KDDE \cite{tomm21kdde}. 
Following \cite{acdc}, we choose DeepLabv2 \cite{chen2017deeplab} as $\mathcal{N}_s$.
As for $\mathcal{N}_t$, we adopt AdaSegNet \cite{tsai2018learning} and FDA \cite{yang2020fda}.

\textbf{Performance metric}. We report \rb{Intersection over Union} (IoU) per class, and mean IoU (mIoU) as the overall performance.

\subsubsection{Results}
\begin{figure}[htbp!]
  \centering
  \includegraphics[width=\columnwidth]{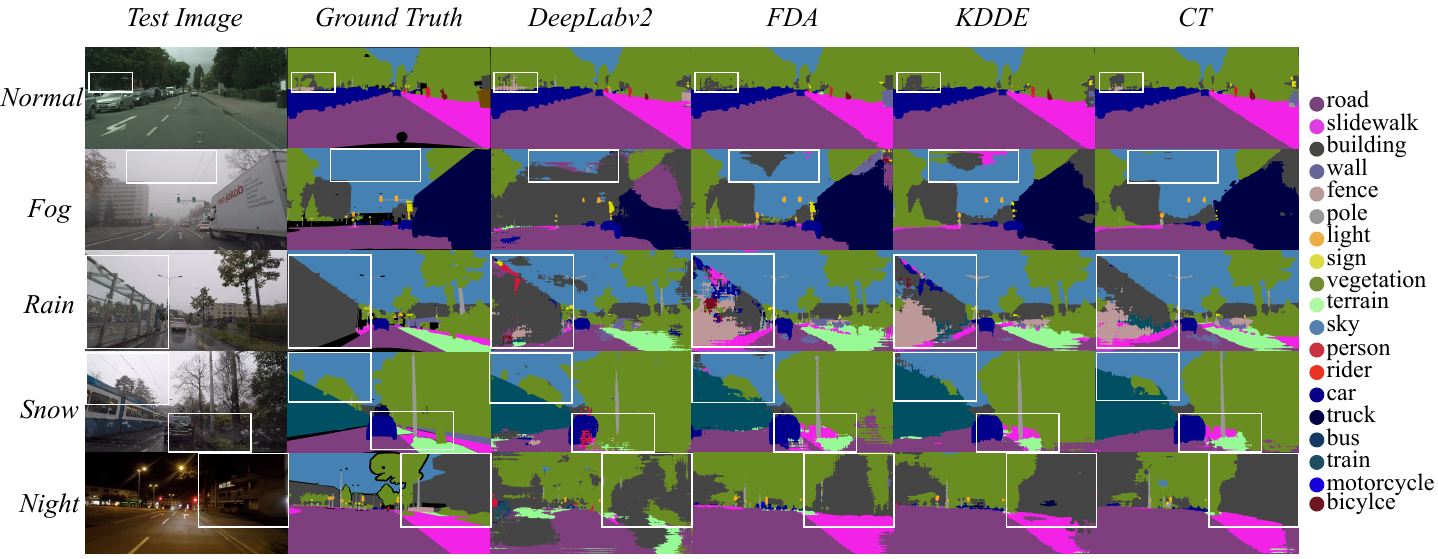}
  \caption{
  \textbf{Qualitative results of driving scene segmentation}. The first row is from $D_s$ (normal condition), while the other rows are from $D_t$ (adverse conditions in the nighttime, fog, snow and rain). Important difference between the results is marked out by white bounding boxes. Best viewed digitally.}
  \label{fig:segmentation}
\end{figure}

As \cref{tab:segmentation_sum} shows, 
the source-domain performance of AdaSegNet and FDA decreases, confirming the necessity of UDE for semantic segmentation. 
Using either AdaSegNet or FDA as its UDA module,  CT restores the source-domain performance. In the adverse conditions, CT reduces misclassification of sky into buildings, see \cref{fig:segmentation}.
Pixel-level classification accuracy is given in 
\cref{tab:seg_ambiguity}. The higher accuracy of \model on the pixels with inconsistent $\mathcal{N}_s$ and $\mathcal{N}_t$ predictions shows its effectiveness in tackling the cross-domain ambiguity.

\begin{table}[!t]
\setlength\tabcolsep{2pt}
\centering
\renewcommand{\arraystretch}{1.1}
\caption{\textbf{Driving scene segmentation}.}
\scalebox{0.9}
{

\begin{tabular}{@{}lrrr@{}}
\hline
\textbf{Method} & $D_s$ & $D_t$ & $D_{s+t}$ \\ \hline
DeepLabv2 as $\mathcal{N}_s$ & \textbf{63.34} & 30.25 & 49.29 \\ 
\midrule
AdaSegNet \cite{tsai2018learning}  & 61.69 & 38.20 & 52.22 \\
FDA \cite{yang2020fda} & 60.93 & 41.11 & 53.33 \\ 
\midrule
\multicolumn{4}{@{}l}{AdaSegNet as $\mathcal{N}_t$:} \\ 
KDDE & 61.98 & 39.47 & 53.55 \\
CT & 62.72 & 40.88 & 54.27 \\ 
\midrule
\multicolumn{4}{@{}l}{FDA as $\mathcal{N}_t$:} \\
KDDE & 62.09 & 41.83 & 54.38 \\
CT & 62.18 & \textbf{42.77} & \textbf{55.14} \\ 
\bottomrule
\end{tabular}

}
\label{tab:segmentation_sum}
\end{table}

\begin{table}[!htbp]
\centering
\renewcommand\arraystretch{1.1}
\caption{\textbf{Pixel-level classification accuracy}.}
\scalebox{0.9}
{
\begin{tabular}{lcc|cc|cc}
\hline
 & \multicolumn{2}{c|}{$D_{s}$} & \multicolumn{2}{c|}{$D_{t}$} & \multicolumn{2}{c}{$D_{s+t}$} \\ \cline{2-7}

\multirow{-2}{*}{\begin{tabular}[c]{@{}l@{}}\textbf{Method}\end{tabular}} & \cellcolor[HTML]{DBF5DB} $=$ & \cellcolor[HTML]{F8CECC}$\neq$ & \cellcolor[HTML]{DBF5DB}$=$ & \cellcolor[HTML]{F8CECC}$\neq$ & \cellcolor[HTML]{DBF5DB}$=$& \cellcolor[HTML]{F8CECC}$\neq$ \\ \hline
DeepLabv2 &  \cellcolor[HTML]{DBF5DB}95.99 & \cellcolor[HTML]{F8CECC}3.06 &
\cellcolor[HTML]{DBF5DB} 86.31 & \cellcolor[HTML]{F8CECC}9.92&
\cellcolor[HTML]{DBF5DB}92.20 & \cellcolor[HTML]{F8CECC}5.74 \\
FDA &  \cellcolor[HTML]{DBF5DB}95.99 & \cellcolor[HTML]{F8CECC}2.55 &
\cellcolor[HTML]{DBF5DB} 86.31 & \cellcolor[HTML]{F8CECC}21.30&
\cellcolor[HTML]{DBF5DB}92.20 & \cellcolor[HTML]{F8CECC}9.89 \\
\model &  \cellcolor[HTML]{DBF5DB}95.29 & \cellcolor[HTML]{F8CECC}3.49 &
\cellcolor[HTML]{DBF5DB}86.66 & \cellcolor[HTML]{F8CECC}26.13 &
\cellcolor[HTML]{DBF5DB}91.91 & \cellcolor[HTML]{F8CECC}12.35 \\ 
\hline
\end{tabular}
}

\label{tab:seg_ambiguity}
\end{table}

\section{Conclusions}
\label{sec:conclusion}
This paper develops Co-Teaching (\model), a new method for unsupervised domain expansion (UDE). Extensive experiments on multi-class image classification 
and driving scene support our conclusions as follows. Due to the existence of cross-domain ambiguity, a domain-specific model is not universally applicable to handle samples from its targeted domain.  \model, with its ability to resolve such ambiguity, provides a unified framework to improve a model's performance on the target domain, and meanwhile maintains mostly its performance on the source domain. For its simplicity and effectiveness, \model is a new baseline for UDE.

\subsubsection{\ackname} This research was supported by National Natural Science Foundation of China (62576348, 62172420) and Beijing Natural Science Foundation (L254039).

\bibliographystyle{splncs04} 
\bibliography{refs}

\end{document}